\documentclass[letterpaper, 10 pt, conference]{ieeeconf}
\IEEEoverridecommandlockouts
\usepackage[T1]{fontenc}
\usepackage{graphics}
\usepackage{amsmath}
\usepackage{amssymb}
\usepackage[usenames, dvipsnames]{xcolor}
\DeclareMathOperator*{\argmax}{arg\,max}
\DeclareMathOperator*{\argmin}{arg\,min}

\usepackage{lipsum}
\usepackage{tikz}
\usepackage{tkz-graph}
\usepackage{subcaption}
\usepackage{float}
\usepackage{soul} 
\usepackage{bm}
\usepackage{cite}

\usepackage{adjustbox}
\usepackage{algorithm}
\usepackage{algpseudocode}

\makeatletter
\let\NAT@parse\undefined
\makeatother

\usepackage{hyperref}
\usepackage[capitalize]{cleveref}

\hypersetup{
  linkcolor  = BrickRed,
  citecolor  = Green,
  urlcolor   = NavyBlue,
  colorlinks = true,
}

\makeatletter
\newcommand{\subalign}[1]{%
  \vcenter{%
    \Let@ \restore@math@cr \default@tag
    \baselineskip\fontdimen10 \scriptfont\tw@
    \advance\baselineskip\fontdimen12 \scriptfont\tw@
    \lineskip\thr@@\fontdimen8 \scriptfont\thr@@
    \lineskiplimit\lineskip
    \ialign{\hfil$\m@th\scriptstyle##$&$\m@th\scriptstyle{}##$\hfil\crcr
      #1\crcr
    }%
  }%
}
\makeatother

\title{\LARGE \bf
Cooperative Control of Mobile Robots with Stackelberg Learning
}

\author{Joewie J. Koh*, Guohui Ding*, Christoffer Heckman, Lijun Chen, Alessandro Roncone%
\thanks{This material is based upon work supported by the National Science Foundation under Grant No. 1646556. The authors also acknowledge support from a ``PyRobot: Democratizing Robotics'' Research Award from Facebook Research. \textit{(Joewie J. Koh and Guohui Ding contributed equally to this work.)}}%
\thanks{All authors are with the Department of Computer Science, University of Colorado Boulder. {\tt\footnotesize \{firstname.lastname\}@colorado.edu}}%
}

\begin{document} 
\bstctlcite{IEEEexample:BSTcontrol}

\maketitle
\thispagestyle{empty}
\pagestyle{empty}


\begin{abstract}
Multi-robot cooperation requires agents to make decisions that are consistent with the shared goal without disregarding action-specific preferences that might arise from asymmetry in  capabilities and individual objectives.
To accomplish this goal, we propose a method named SLiCC: \underline{S}tackelberg \underline{L}earning \underline{i}n \underline{C}ooperative \underline{C}ontrol.
SLiCC models the problem as a partially observable stochastic game composed of Stackelberg bimatrix games, and uses deep reinforcement learning to obtain the payoff matrices associated with these games.
Appropriate cooperative actions are then selected with the derived Stackelberg equilibria.
Using a bi-robot cooperative object transportation problem, we validate the performance of SLiCC against centralized multi-agent Q-learning and demonstrate that SLiCC achieves better combined utility.
\end{abstract}


\section{Introduction}\label{sect:introduction}

\begin{figure}
	\centering
    \begin{subfigure}[b]{0.22\textwidth}
    	\centering
        \includegraphics[width=1.0\linewidth]{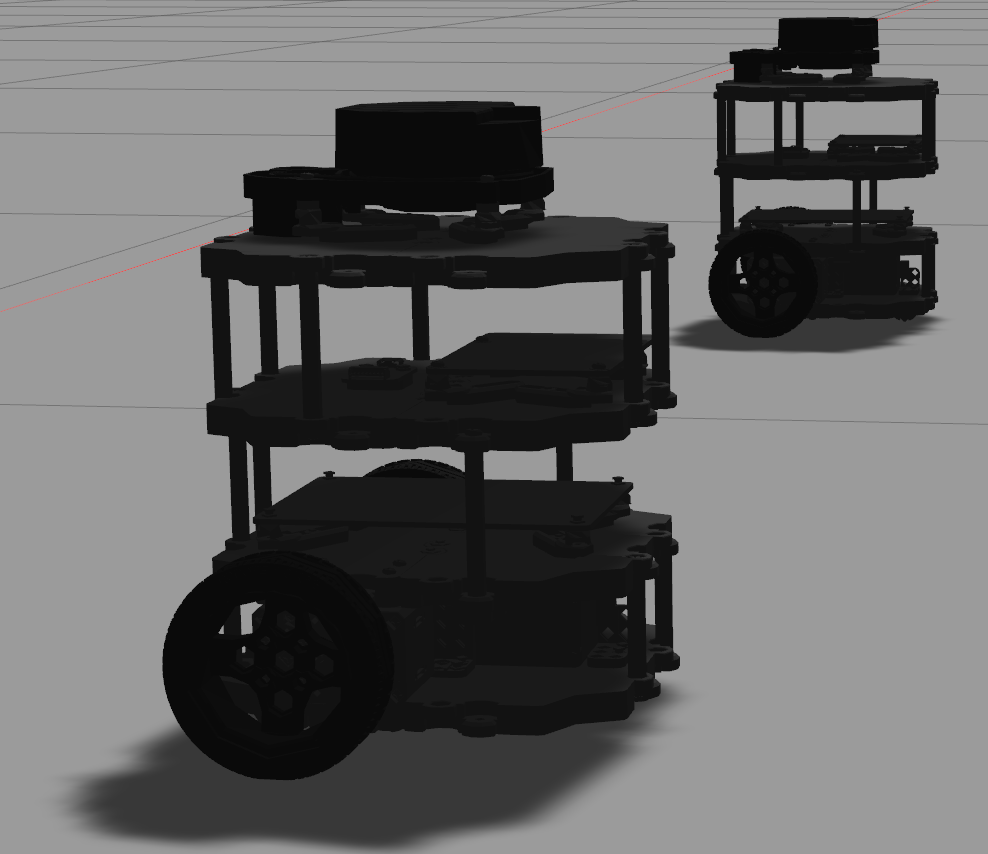}
        \caption{Two TurtleBot3 Burger mobile robots in the Gazebo simulator.}\label{fg:work_scenario_turtlebot3}
    \end{subfigure}
    \hspace{0.25cm}
    \begin{subfigure}[b]{0.22\textwidth}
    	\centering
        \includegraphics[width=1.0\linewidth]{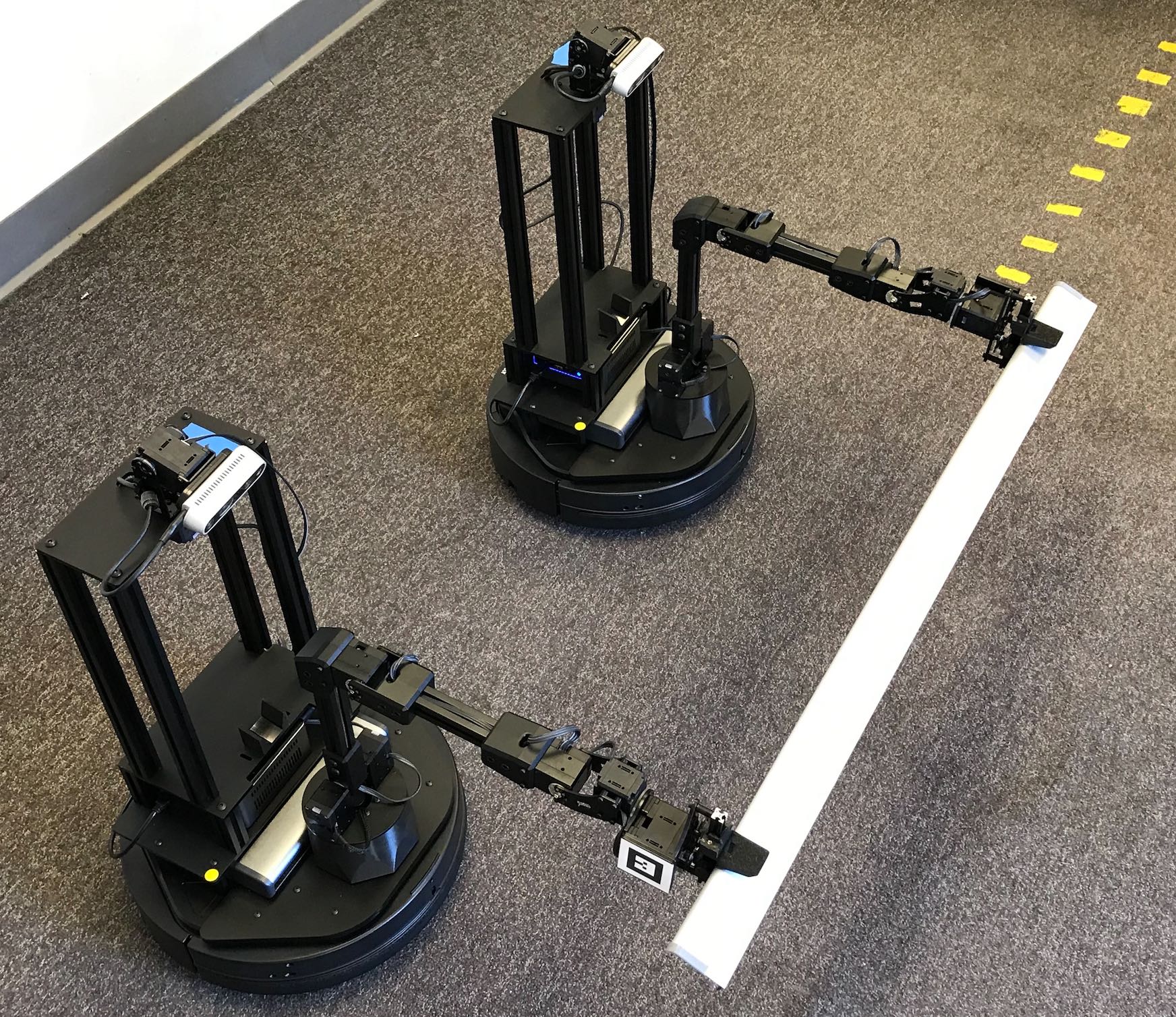}
        \caption{Two LoCoBot mobile robots cooperatively transport an object.}\label{fg:work_scenario_locobot}
    \end{subfigure}
    \caption{SLiCC is a novel method designed for cooperative control of bi-robot systems. In this work, we evaluate the performance of SLiCC using a pair of simulated robots and another pair of real robots. The two robots were selected for their different physical dynamics.}\label{fg:work_scenario}
\end{figure}

Robotics at large has been improving at a rapid pace, and this has resulted in increased demand in applications ranging from manufacturing \cite{schneier2015literature}, to warehousing \cite{bogue2016growth}, to human-populated environments \cite{ajoudani2018progress}.
However, despite the clear potential for distributed controllers that leverage cooperation between multiple mobile robots (e.g., the cooperative transport of large or heavy objects, see \cref{fg:work_scenario_locobot}), the vast majority of existing techniques are either limited to single-robot operation or require that each robot perceive the complete state of the environment \cite{nguyen2020deep}.
Interestingly enough, model-free learning-based methods present a promising alternative to traditional model-based control, in that they are less reliant on domain knowledge such as kinematic and dynamic modeling of the system, and that they scale more naturally with the number of agents.

Notably, reinforcement learning (RL) shows promise for controlling robotic systems in unstructured environments as it enables robots to discover useful behaviors simply by interacting with the environment \cite{kober2013reinforcement}.
However, research in the field is still at an early stage, and state-of-the-art methods are often limited by issues such as partial observability, particularly in multi-agent settings \cite{hernandez2019survey}.
Learning cooperative control of multi-robot systems to achieve common objectives can be difficult in practice due to agents having asymmetric embodiments and capabilities; for example, robot participants can be limited by perceptual \cite{pereira2003decentralized}, communicative \cite{burgard2005coordinated}, locomotive \cite{miki2019uav}, or computational \cite{ichnowski2020economic} constraints to different extents.
Additionally, agent-specific preferences in a cooperative setting can inadvertently interfere with the successful accomplishment of the common objective, leading to unstable learning in na\"{i}ve multi-agent adaptations of single-agent RL methods.
To help agents make appropriate decisions in these circumstances, it is critical to utilize  well-assigned cooperation responsibilities without disregarding agent-specific preferences---this motivates the use of negotiated decision-making processes.

In this paper, we introduce SLiCC: \underline{S}tackelberg \underline{L}earning \underline{i}n \underline{C}ooperative \underline{C}ontrol, which uses a novel prosocial--introspective framework to achieve a common goal in a bi-agent system.
The proposed framework allows for agents to have different observation scopes, with prosocial and introspective behaviors assigned to agents based on the completeness of their state perception.

The main idea behind the prosocial--introspective framework is to provide a link between perception asymmetry and agent utility, in the context of cooperative control.
More specifically, in order to successfully achieve a shared goal (e.g., two robots tasked with jointly carrying an object), the efficient exploitation of inter-agent interaction dynamics and consequent allocation of learning responsibilities can compensate for partial observability that arises as a consequence of perception asymmetry.

Deriving these interaction dynamics requires additional perceptual capabilities, either in terms of perceiving the complete state (e.g., the positions of both robots) or interaction forces (e.g., tension or compression on jointly-carried objects).
Therefore, the utility of the agent with such additional perceptual capabilities should be dependent on both agents' decisions; we refer to this agent as being \textit{prosocial}---its behavior is largely motivated by concerns about obeying constraints of the common goal.
On the other hand, the agent with only partial state perception has utility that is only dependent on its own decisions; we therefore refer to this as the \textit{introspective} agent---its attention is exclusively on its own state and utility.

More specifically, SLiCC models the prosocial--introspective cooperation problem as a partially observable stochastic game (POSG) composed of Stackelberg bimatrix games.
Decomposing a POSG into Stackelberg bimatrix games allows the use of the Stackelberg equilibrium at each decision step to approximate the POSG equilibrium.
With the insight that an agent's payoff function is equivalent to their Q-value function, we use deep RL to learn payoff matrices, capitalizing on the function approximation capabilities of deep learning to do so even in settings with continuous state spaces.
The Stackelberg equilibrium can then be derived from the agents' payoff matrices and used to inform the agents' actions, serving as a mechanism for negotiated decision-making.

Our contributions are as follows: 
\begin{enumerate}
    \item Introduce SLiCC, a method for cooperative control of bi-agent systems in partially observable settings based on an asymmetric prosocial--introspective cooperation framework that links state perception with agents' decision-making strategies.
    \item Provide an improved Stackelberg game--based architecture to enhance the agents' policy learning capabilities under POSG settings.
    \item Demonstrate that a SLiCC policy learned in simulation can be used to control real robots without additional training.
\end{enumerate}

The rest of the paper has the following structure.
To begin with, \cref{sect:related_work} presents an overview of the relevant literature.
Following a description of the problem setting in \cref{sect:problem_setting}, we elucidate the proposed SLiCC method in \cref{sect:stackelberg_learning}: we begin with a brief treatment of POSGs, and then explain how we bridge the POSG setting and our prosocial--introspective framework using the Stackelberg equilibrium.
Next, \cref{sect:comparison} discusses key differences of SLiCC compared with other learning paradigms.
The advantages of SLiCC detailed in this section are consequential in real-world applications of cooperative control.
\cref{sect:experiments_and_evaluations} follows with our experimental setup and results.
Besides evaluating SLiCC in simulation, we also validate our proposed method with real robots.
Finally, \cref{sect:conclusion} concludes the paper and provides an inventory of future directions.


\section{Related Work}\label{sect:related_work}

Multi-robot cooperative control has traditionally been pursued from the perspective of control theory \cite{khatib1996vehicle, ogren2002control, feddema2002decentralized}. 
Indeed, multi-agent scenarios present peculiar challenges and constraints for RL-based solutions \cite{hernandez2019survey}, which has limited their relevance for the field despite recent successes of deep RL.
Methods such as policy search enable robots to learn complex real-world skills in single-agent settings (e.g., door opening \cite{yahya2017collective} and map-less navigation \cite{tai2017virtual}).
Unfortunately, it is difficult to extend these methods to multi-robot applications.
Doing so with a centralized learning paradigm leads to exponentially increasing state and action spaces \cite{fitch2008million}, and consequently, infeasible computational needs.
Conversely, agent decentralization--based RL approaches are not guaranteed to result in stable policies \cite{tan1993multi}.
However, there is a promising line of research with game-theoretic RL: policy updates can be based on stochastic game equilibria, with the goal of improving state-value function estimates and reducing learning instability.
This has been demonstrated as an effective solution that incorporates agent--agent interaction in multi-agent RL \cite{littman1994markov, hu2003nash, ding2020distributed}, and these approaches are seeing increasing interest across different communities \cite{nikolaidis2017game, apostolopoulos2018demand, ding2018gametheoretic}.

Nonetheless, this direction is not without limitations.
Stochastic games typically assume that all agents have complete observations \cite{littman1994markov, hu2003nash, ding2020distributed}, which is a limiting constraint for some settings---and in particular for multi-agent applications.
This point provides an opportune segue into partially observable stochastic games (POSGs), which arise naturally from asymmetry in agent-specific preferences, cooperation responsibilities, or observation scopes.
In previous work, the POSG learning process has been approximated with a series of Bayesian games and common payoffs, using belief spaces to compensate for missing observations \cite{emerymontemerlo2004approximate}.
Methods like Joint Equilibrium--based Search for Policies (JESP) \cite{nair2003taming} have also been used to alternately optimize the agents' objective.
Improved flexibility in the decentralized learning process has furthermore been achieved via modeling an opponent's policy through recursive reasoning \cite{wen2019probabilistic}. 

These works motivate us to further consider how to coordinate asymmetric agent roles engendered by robots with different perceptual capabilities.
Previous research in this area attempted to do so by incorporating Stackelberg games into RL techniques \cite{kononen2004asymmetric, laumonier2005multiagent, shi2020learning}.
Similar approaches have been applied in human--robot interaction \cite{nikolaidis2017game} and smart microgrids \cite{wang2016reinforcement}, but they dictate a common state space for all agents.
Inspired by recent work applying Stackelberg games to partially observable scenarios \cite{alcantarajimenez2020repeated}, we present a novel improvement for robot learning by integrating Stackelberg games with POSGs and reducing the prerequisite of complete state observations through our prosocial--introspective framework.
In order to optimize the performance of game-theoretic RL approaches, we investigate agent roles in a multi-robot cooperation problem with asymmetric information.
In the rest of the paper, we demonstrate the potential of game-theoretic RL approaches with SLiCC as an exemplar.


\section{Problem Setting and Generalizability}\label{sect:problem_setting}

For convenience, in this work we consider a bi-agent system composed of two mobile robots---although the proposed method can generally be applied to a variety of multi-robot scenarios.
The state of each robot is denoted with a tuple $(x^k, y^k, \theta^k, v^k)$, where $x^k$ and $y^k$ are the robot's coordinates in the $2$-dimensional plane, and $\theta^k$ is the robot's orientation about the $z$-axis.
Furthermore, the magnitude of the agent's linear velocity $v^k$ in the $2$-dimensional plane is given by the $L_2$-norm of the time derivatives of $x^k$ and $y^k$.

Accordingly, with $a^{\omega, k}_{t}$ and $a^{v, k}_{t}$ being the actionable increments for angular velocity and linear velocity respectively, the robots' dynamics in discrete space (which is the same as that which is used by the RL agents) is given by:
\begin{align}
    & x^k_{t + 1} = x^k_{t} + v^k_{t} \cos(\theta^k_t) \Delta t + \epsilon_{x^k}
    , \label{eq:dynamics_x} \\
    & y^k_{t + 1} = y^k_{t} + v^k_{t} \sin(\theta^k_t) \Delta t + \epsilon_{y^k}
    , \label{eq:dynamics_y} \\
    & \theta^k_{t + 1} = \theta^k_{t} + a^{\omega, k}_{t} \Delta t + \epsilon_{\theta^k}
    , \label{eq:dynamics_theta} \\
    & v^k_{t + 1} = v^k_{t} + a^{v, k}_{t} + \epsilon_{v^k}
    , \label{eq:dynamics_v}
    \end{align}
where $\epsilon_{(\cdot)}$ denotes environment-induced noise (e.g., wheelspin) and errors incurred from the discretization of the dynamics.
The robots are given the objective of cooperatively transporting an object that is beyond the capabilities of a single robot to carry (see \cref{fg:work_scenario_locobot}).

We selected this particular task because it introduces dynamical constraints that are not explicitly known by all agents, and it is furthermore a convenient setting to investigate perception asymmetry.
In our setting, the first robot (i.e., the prosocial agent) can perceive both robots' state tuples, but the second robot (i.e., the introspective agent) can only perceive its own state tuple.
We also assume that the prosocial agent can receive information from the introspective agent at each decision step.

We elected to demonstrate the merits of our prosocial--introspective framework with this simple bi-agent system as it is representative of imbalanced agent roles in a multi-agent setting.
However, it is worthwhile to note that SLiCC is applicable too in systems with more than two agents: we can easily add more introspective agents without loss of generality.


\section{Stackelberg Learning in Cooperative Control}\label{sect:stackelberg_learning}

\begin{figure}
	\centering
    \includegraphics[width=\linewidth]{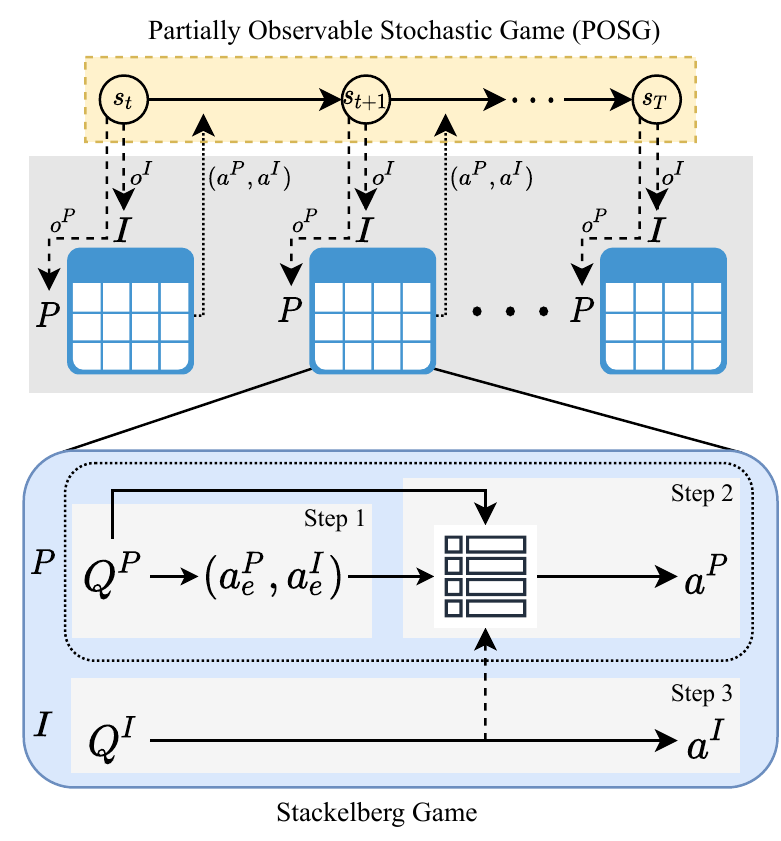}
    \caption{Outline of the SLiCC algorithm. ($P$: Prosocial Agent, $I$: Introspective Agent.) Each decision step $t$ of the POSG corresponds to a Stackelberg game (see \cref{sect:stackelberg_game}). The Q-table of each agent is estimated based on its observations. Step 1: Agent $P$ determines the expected action pair $(a^P_e, a^I_e)$ based on its own Q-table; Step 2: Agent $P$ derives the actual action $a^P$; Step 3: Agent $I$ determines its actual action $a^I$.}\label{fg:slicc_outline}
\end{figure}

In this section, we present the main components of SLiCC (see \cref{fg:slicc_outline} for an outline of the method).
As previously mentioned, we use the POSG model to characterize the policy learning process due to the difference in observation scopes between the two agents.
The prosocial--introspective framework allows us to relate this difference to agent utility, while step-wise Stackelberg equilibria are used to approximate the POSG equilibrium.
Finally, we describe how deep Q-networks can be used to estimate the agents' Q-values.

\subsection{Partially Observable Stochastic Game (POSG)}

A POSG can be represented by a tuple $\mathcal{G}$, where
\begin{eqnarray*}
    \mathcal{G} = \left\{
        \mathcal{N},
        S,
        \left( O^k \right)_{k \in \mathcal{N}},
        \left( A^k \right)_{k \in \mathcal{N}},
        T,
        \left( r^k \right)_{k \in \mathcal{N}},
        \left( \pi^k \right)_{k \in \mathcal{N}}
    \right\}.
\end{eqnarray*}
The elements of $\mathcal{G}$ are defined as follows:
\begin{itemize}
    \item $\mathcal{N} = \left\{ 1, \dotsc, n \right\}$ is the set of agents;
    \item $S$ is the set of states, common to all agents;
    \item $O^k \subseteq S$ is the observation space for agent $k$;
    \item $A^k$ is the set of actions available to agent $k$;
    \item $T \colon S \times A^1 \times \cdots \times A^n \rightarrow PD(S)$ is the state transition function, with $PD(\cdot)$ being a probability distribution, and with the transition dynamics common to all agents;
    \item $r^k \colon O^k \times A^1 \times \cdots \times A^n \rightarrow \mathbb{R}$ is the reward function for agent $k$; and
    \item $\pi^k \colon O^k \rightarrow PD(A^k)$ is the policy of agent $k$.
\end{itemize}
As detailed in \cref{sect:problem_setting}, in this work we consider a bi-agent system consisting of a prosocial agent and an introspective agent, i.e., $\mathcal{N} = \{ P, I \}$.
Therefore, $\forall o \in O^k$, the Stackelberg equilibrium $(\pi^{P^*}, \pi^{I^*})$ \cite{kononen2004asymmetric} satisfies \cref{eq:equilibrium_prosocial,eq:equilibrium_introspective}, for all $\pi^{P}$ and $\pi^{I}$ respectively:
\begin{align}
    V^{P} \left( o \middle| \pi^{P^*}, \pi^{I^*} \right) & \geq
    	V^{P} \left( o \middle| \pi^{P}, BR(V^I, \pi^{P}) \right)
    	, \label{eq:equilibrium_prosocial} \\
    V^{I} \left( o \middle| \pi^{P^*}, \pi^{I^*} \right) & \geq
    	V^{I} \left( o \middle| \pi^{P^*}, \pi^{I} \right)
    	. \label{eq:equilibrium_introspective}
\end{align}

Here, $BR(V^I, \pi^{P}) = \argmax_{u^I \in A^I} V^{I} \left( o \middle| \pi^{P}, u^{I} \right)$ represents the best response of Agent~$I$ given Agent~$P$'s policy.
For $k \in \mathcal{N}$, $V^k(\cdot)$ refers to the discounted episode reward with observation $o$ at time $t$,
\begin{eqnarray}
    V^k(o) = \mathbf{E} \left[
        \sum_{i = 0}^{T-t} \gamma^{i} r^k_{t+i} \ \middle| \ o_t = o, (\pi^P, \pi^{I})
    \right]
    .
\end{eqnarray}
\cref{eq:equilibrium_prosocial,eq:equilibrium_introspective} essentially state that, when agent $k$ follows an equilibrium policy $\pi^{k^*}$, the discounted episode reward attained is guaranteed to be greater than or equal to that attained with any other policy.
Agent $P$'s decision is also related to the best response of Agent $I$.
In our problem, we only consider stationary policies $\left( \pi^k = \pi^k(s^1), \dotsc \right)$.

\subsection{Prosocial--Introspective Framework}

Under the POSG setting, each agent makes decisions based on its observations.
The prosocial agent (Agent $P$) has complete observation of the system (i.e. perfect perception), but the introspective agent (Agent $I$) only observes its own state.
Specifically for our problem setting, the state of agent $k$ is $s^k = (x^k, y^k, \theta^k, v^k)$.
Thus, the observations of the two agents are $o^P = (s^P, s^I)$ and $o^I = s^I$.
The reward function of each agent depends on its observation and action spaces: $r^k : O^k \times A^k \rightarrow \mathbb{R}$.

\subsection{Stackelberg Game}\label{sect:stackelberg_game}

We use a Stackelberg game formulation as the intermediary between our prosocial--introspective framework and the POSG model, as shown in \cref{fg:slicc_outline}.
We express each decision step of the POSG as a partially observed bimatrix game $\mathcal{G}^{SG}$, where
\begin{align}
    \mathcal{G}^{SG}(s) = \left\{
         \mathcal{N},
         (Q^k(\cdot))_{k \in \mathcal{N}},
         (\pi^k)_{k \in \mathcal{N}},
         s
    \right\}.
\end{align}
Here, $Q^k$ refers to the payoff matrix of Agent $k$.
Note that an agent's payoff matrix is also its Q-table: respectively, $Q^P \left( o^P, a^P, a^I \right)$ and $Q^I \left( o^I, a^I \right)$ are the Q-values for the Agents $P$ and $I$ when observing $o^P$ and $o^I$ \cite{laumonier2005multiagent}.
To derive a Stackelberg equilibrium solution and actions at each decision step $t$, the agents use the following steps:

\begin{enumerate}
    \item Agent $P$, using its own $Q^P$, determines the expected action pair that maximizes $Q^P$:
    \begin{align}
        (a^P_e, a^I_e) = \argmax_{\subalign{u^P &\in A^P \\ u^I &\in A^I}} Q^P(o^P_t, u^P, u^I)
        . \label{eq:slicc_update_1}
    \end{align}
    \item Agent $P$ receives $Q^I$ from Agent $I$, and derives an actual action $a^P$ by minimizing the difference between the Q-values obtained from the actual and expected action pairs:
    \begin{align}
        a^P =  \argmin_{u^P \in A^P} | & Q^P ( o^P_t, u^P, \argmax_{u^I \in A^I}Q^I (o^I_t, u^I) ) \nonumber \\
        & - Q^P(o^P_t, a^P_e, a^I_e) |
        . \label{eq:slicc_update_2}
    \end{align}
    \item Agent $I$, using its own $Q^I$, determines its actual action $a^I$  that maximizes $Q^I$:
    \begin{align}
        a^I = \argmax_{u^I \in A^I} Q^I(o^I, u^I)
        . \label{eq:slicc_update_3}
    \end{align}
\end{enumerate}

\subsection{Algorithm Overview}

\begin{algorithm*}
	\caption{Stackelberg Learning in Cooperative Control (SLiCC)}\label{al:slicc}
	\begin{algorithmic}[1]
    	\State Initialize neural networks $Q^k(\cdot; \theta^k)$ with corresponding output size ($P: |A^P| \times |A^I|$, $I: |A^I|$) and replay buffer $\mathcal{D}$.
		\For {episode $1 : M$}
        	\State Initialize state $s_0$ for all agents.
			\For {$t=1,\dotsc, T$}
                \State $P$ derives expected actions $(a^P_e, a^I_e) = \argmax_{u^P, u^I} Q^P(o^P_t, u^P, u^I; \theta^P)$.
                \State $P$ and $I$ simultaneously $\epsilon$-greedily execute $a^P$ and $a^I$ respectively:
                \begin{align}
                    \begin{cases}
                        a^P = \argmin_{u^P} | Q^P(o^P_t, u^P, \argmax_{u^I}Q^I(o^I_t, u^I)) - Q^P(o^P_t, a^P_e, a^I_e) |, \\
                        a^I = \argmax_{u^I} Q^I(o^I, u^I).
                    \end{cases} \nonumber
                \end{align}
                \State Both agents add their corresponding transition tuple $\tau^k_t = (o^k_t, a^k_t, r^k_t, o^k_{t + 1})$ to the replay buffer $\mathcal{D}$.
                \For {$(\tau^P, \tau^I)_i$ in the mini-batch sample from $\mathcal{D}$}
                    \State $P$ calculates target as follows:
                    $\ell^P_i =
                        \begin{cases}
                            r^P, & \text{if episode terminates;} \\
                            r^P + \gamma \max_{u^P} Q^P(o^{P'}, u^P, F^I_t(o^{I'}); \theta^P), & \text{otherwise (see \cref{eq:argmax_helper}).}
                        \end{cases}
                    $
                    \State $I$ calculates target as follows:\hspace{1.4mm}
                    $\ell^I_i =
                        \begin{cases}
                            r^I, &\hspace{17.2mm}\text{if episode terminates;} \\
                            r^I + \gamma \max_{u^I} Q^I(o^{I'}, u^I; \theta^I), &\hspace{17.2mm}\text{otherwise.}
                        \end{cases}
                    $
                    \State Update $\theta^k$ by minimizing the cost $(\ell^k_i - Q^k((o^k, a^k)_i; \theta^k))^2$.
                \EndFor 
			\EndFor
		\EndFor
	\end{algorithmic}
\end{algorithm*}

The SLiCC method is described in \cref{al:slicc}.
First, both Agent $P$ and Agent $I$ initialize their respective neural networks to approximate $Q^P$ and $Q^I$ respectively.
Then, during each episode, the agents use the Stackelberg equilibrium to guide their action choices, i.e. using \cref{eq:slicc_update_2,eq:slicc_update_3}.
Finally, the agents update their Q-value functions:
\begin{align}
    & Q^P_{t+1} \left(
        o^P_t, a^P, a^I
    \right)
    = (1 - \alpha_t) \cdot Q^P_t \left( o^P_t, a^P, a^I \right) \nonumber \\
    & \quad + \alpha_t \left[
        r^P_t + \gamma \max_{u^P \in A^P} Q^P_t \left(o^P_{t + 1}, u^P, F^I_t(o^I_{t + 1}) \right)
    \right]
    , \label{eq:q_update_prosocial} \\
    & Q^I_{t+1} \left(
        o^I_t, a^I
    \right)
    = (1 - \alpha_t) \cdot Q^I_t \left( o^I_t, a^I \right) \nonumber \\
    & \quad + \alpha_t \left[
        r^I_t + \gamma \max_{u^I \in A^I} Q^I_t \left(o^I_{t + 1}, u^I \right)
    \right]
    . \label{eq:q_update_introspective}
\end{align}
For conciseness of notation in \cref{eq:q_update_prosocial}, we have defined:
\begin{align}
    F^I_t(x) = \argmax_{u^I \in A^I} Q^I_t(x, u^I)
    . \label{eq:argmax_helper}
\end{align}

From \cref{eq:q_update_introspective}, we see that the Agent $I$'s policy learning does not depend on Agent $P$'s actions.
Conversely, \cref{eq:q_update_prosocial} shows that Agent $P$ learns to adapt to Agent $I$ during its Q-value updates.
This reinforces our interpretation of the prosocial--introspective framework: the introspective agent only focuses on its own state and utility, while the prosocial agent ensures the common objective is fulfilled by making adaptive decisions and reconciling actions taken by the introspective agent.
Ultimately, the approach allows us to mediate the agents' perception asymmetry by means of appropriately assigning cooperation responsibilities.


\section{Comparison with Other Learning Paradigms}\label{sect:comparison}

We now discuss several key aspects in which SLiCC differs from other learning paradigms.
We address them with regard to two main themes: perceptual or communicative requirements, and time-varying agent-specific preferences.
These have important implications when considering the design specification and adaptability of a multi-robot system.

\subsection{Perceptual or Communicative Requirements}

SLiCC places less emphasis on the capabilities of individual agents compared to other frameworks.
As previously mentioned, SLiCC does not require all agents to have complete state observations.
This means that we can avoid equipping introspective agents with the full suite of sensors typically required for the task.
Alternatively, we can reduce the required communication bandwidth as agents no longer have to be in constant two-way communications with each other to share state information.
Furthermore, SLiCC operates with full effectiveness using one-way communications: introspective agents do not need to receive information from any other agent.

\subsection{Time-Varying Agent-Specific Preferences}

We consider here a situation that can further clarify some advantages of the proposed SLiCC method.
Imagine a scenario where an agent suffers a mechanical malfunction, therefore impeding its ability to carry out a certain action as was originally determined at the design step of the problem.
In such a scenario, it would be beneficial for the agent to modify its policy to accommodate its diminished capabilities.

\subsubsection{Independent Learning}

With independent learning, each agent disregards the existence of the other agents, subsuming them within the environment.
This approach requires no inter-agent communication, and agents seek to maximize their individual utilities.
While independent learning might allow to learn satisfactory policies in some multi-agent settings, it is expected to fail in this scenario given the additional agent--agent dynamics in the environment.

\subsubsection{Centralized Learning}\label{sect:centralized_learning}

Centralized learning utilizes a single policy to jointly control all agents, managing the multi-agent system as if it were a single complex agent.
This approach is, in theory, capable of adapting to this scenario in two ways.
First, the learning process can be designed a priori to accommodate such an eventuality, by introducing agent-specific preferences explicitly in the state space; however, doing so for all agents quickly leads to combinatorial explosion.
Second, the centralized policy can adapt to the malfunctioning robot's modified behavior and action space through learning.
Unfortunately, since the centralized policy's action space is the Cartesian product of all agents' action spaces, it will take a long time to converge to the new policy.

\subsubsection{SLiCC}

Introspective agents in SLiCC are able to implicitly convey their updated agent-specific preferences in the form of $Q^I$.
Following a change in its agent-specific preferences, an introspective agent can quickly update its $Q^I$ since the dimensionality of $Q^I$ is simply equal to the dimensionality of its action space. This stands in direct contrast to the update situation with centralized learning described above.
Upon receiving an updated $Q^I$, Agent $P$ can immediately accommodate the changes in Agent $I$'s agent-specific preferences (see \cref{eq:slicc_update_2}).


\section{Experiments and Evaluations}\label{sect:experiments_and_evaluations}

In this section, we report on the results of our evaluation of SLiCC with two different reward prototypes.
We use SLiCC to learn a policy for the cooperative transport task in the Gazebo robotics simulator with a pair of TurtleBot3 Burger mobile robots (\cref{fg:turtlebot3}).
As a baseline for comparison, we also learn a centralized policy (see \cref{sect:centralized_learning}) with a deep Q-network \cite{mnih2015humanlevel}, referred to hereinafter as centralized Q-learning.
Our code is publicly available on GitHub\footnote{\url{https://github.com/HIRO-group/SLiCC}}, and a summary video of our experiments can be accessed at \url{https://youtu.be/NnuhFeVTcOw}.

\subsection{Reward Structure}

In our problem, the different characteristics of the agents provide an advantageous scenario to design reward functions to cover multiple learning responsibilities.
Specifically, one of the main advantages for our game-theoretic RL approach is the inclusion of agent-specific preferences \cite{hu2003nash}, denoted as $r_{ap}$.
This component characterizes agent-based concerns, preferences, or desires, which are not affected by the other agents or the environment.
In addition, a multi-agent system also needs to carefully balance agent--agent coordination with the achievement of the common goal---which we hereinafter refer to as $r_{int}$ and $r_{goal}$ respectively.

Considering that the prosocial agent has a broader observation scope, it can respond to the introspective agent's possibly unexpected behaviors, without disregarding its agent-specific preferences or the common goal. We have the following reward prototypes:
\begin{enumerate}
    \item $\mathbf{RP_\alpha}$: The prosocial agent incorporates the global goal, agent--agent interaction, and its agent-specific preferences. The introspective agent focuses on the global goal and its agent-specific preferences.
    \begin{align}
        & r^P_t(o^P_t, a^P_t) = r^P_{goal} + r_{int} + r^P_{ap}, \\
        & r^I_t(o^I_t, a^I_t) = r^I_{goal} + r^I_{ap}.
    \end{align}
    \item $\mathbf{RP_\beta}$: The prosocial agent incorporates agent--agent interaction and its agent-specific preferences. The introspective agent focuses on the global goal and its agent-specific preferences.
    \begin{align}
        & r^P_t(o^P_t, a^P_t) = r_{int} + r^P_{ap}, \\
        & r^I_t(o^I_t, a^I_t) = r^I_{goal} + r^I_{ap}.
    \end{align}
\end{enumerate}
In these reward prototypes, $r_{int}$ evaluates the interaction between $P$ and $I$: a higher value implies that the two robots are staying in formation, as is required for cooperative transport.
This term does not contain information about the global target.
Meanwhile, $r_{goal}$ quantifies the discrepancy between the current state and the expected goal state.
Since the two agents cooperatively complete one task, the global goal can be defined based on individual states.
Finally, $r^k_{ap}$ represents the agents' preference for smooth consecutive actions: drastic changes in robot commands might result in undesirable behaviors (e.g., loss of traction).

$\mathbf{RP_\alpha}$ and $\mathbf{RP_\beta}$ present two different cooperation plans of the prosocial agent.
Compared to $\mathbf{RP_\alpha}$, $\mathbf{RP_\beta}$ reduces the prosocial agent's duty towards accomplishing the common goal and allows it to focus on how to negotiate its agent-specific preferences and the agent--agent interactions, while trusting that the limited information available to the introspective agent will be enough to accomplish the task.
Although the consideration of $r^k_{ap}$ introduces additional complexity in the reward landscape and thus complicates the policy learning process, the equilibrium-based policy updates of SLiCC provide flexibility to make appropriate decisions.

In our experiment, the target state is $s^{target} = (x^{target}, y^{target}, v^{target}, \theta^{target})$.
Based on the problem setting and our testbed, we define $r_{ap} = (r^P_{ap}, r^I_{ap})$, $r_{int}$, and $r_{goal} = (r^P_{goal}, r^I_{goal})$ as follows:
\begin{align}
    & r_{int} = -\| \| (x^P_t, y^P_t) - (x^I_t, y^I_t) \| - \sigma \|, \\
    & r^k_{goal} = - \| (v, \theta)^k_t - (v, \theta)^{target} \|, \\
    & r^k_{ap} = \begin{cases}
                    \mu^{upper}, ~~\mbox{if}~~ \| a^{v, k}_{t} - a^{v, k}_{t - 1} \| \leq \zeta, \\
                    -\mu^{lower}, ~~\mbox{else}.
                 \end{cases}
\end{align}
Here $\mu^{upper} > 0$, $\mu^{lower} > 0$, and $\zeta > 0$.
The desired distance between the two agents is denoted as $\sigma$.

\subsection{Training Architecture}

\begin{figure}
	\centering
    \begin{subfigure}[b]{0.45\textwidth}
    	\centering
        \includegraphics[width=\linewidth]{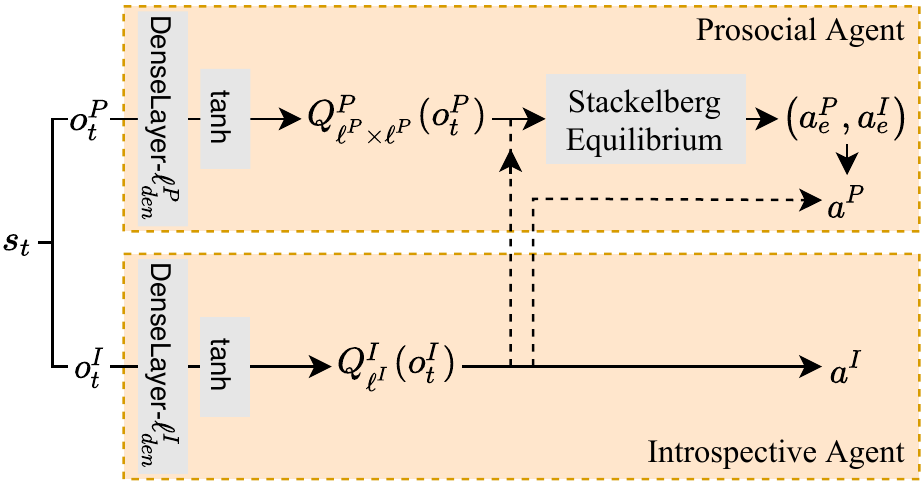}
        \caption{SLiCC: The prosocial and introspective agents use one neural network each to approximate their Q-function. Each network has a single hidden dense layer of size $\ell^P_{den} = 1024$ with a $tanh$ activation function. The networks have different output dimensionality: $\ell^P \times \ell^P$ for agent $P$, and $\ell^I$ for agent $I$. In our experiments, $\ell^P = \ell^I = 9$. The actions are selected as per \cref{eq:slicc_update_1,eq:slicc_update_2,eq:slicc_update_3}.}\label{fg:network-slicc}
    \end{subfigure} \\
    \vspace{0.3cm}
    \begin{subfigure}[b]{0.45\textwidth}
    	\centering
        \includegraphics[width=\linewidth]{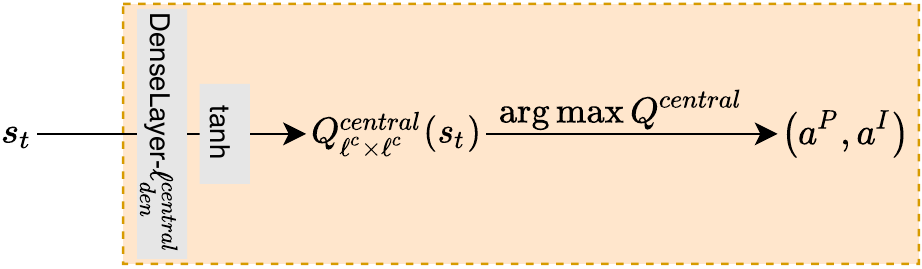}
        \caption{Centralized Q-learning: The centralized learning network has a single hidden dense layer of size $\ell^{central}_{den} = 1024$ with a $tanh$ activation function. The output dimensionality is $\ell^c \times \ell^c$ is the number of output size, with $\ell^c = 9$. The action pair $(a^P, a^I)$ is that which maximizes the Q-value.}\label{fg:network-centralized}
    \end{subfigure}
    \caption{The neural network structures for our experiments.}\label{fg:network_structure}
\end{figure}

The neural network structure for \cref{al:slicc} is shown in \cref{fg:network-slicc}.
Each RL agent has a fully connected neural network with a single hidden dense layer containing $\ell^k_{den}$ neurons.
The input layer of each network has the same dimensionality as its corresponding agent's observation $o^k$.
The output size of each neural network has the same dimensionality as the corresponding agent's Q-table ($P: \ell^P \times \ell^P, I: \ell^I$).
Based on the requirements of our problem setting and testbed, $\ell^P = \ell^I = 9$ and $\ell^k_{den} = 1024$.
Relevant hyperparameters used in learning are: $\gamma = 0.95$, and $|\mathcal{D}_{\min}| = 64$.
The parameters used in the reward are: $\mu^{upper} = 0.05$, $\mu^{lower} = 0.02$, and $\zeta = 0.03$.
These values were selected via hyperparameter search, but our empirical experience suggests that SLiCC is relatively robust to reasonable changes in hyperparameters.

The action space is $\mathbf{A}^{type} = \{ 0, -2 \Delta a^v, -\Delta a^v, \Delta a^v, \\ 2 \Delta a^v, -2 \Delta a^{\theta}, -\Delta a^{\theta}, \Delta a^{\theta}, 2 \Delta a^{\theta} \}$.
For each agent, the actual action at each decision step is an angular velocity change ($a^{\omega, k}_{t} \in \mathbf{A}^{type}$) or a linear velocity increment ($a^{v, k}_{t} \in \mathbf{A}^{type}$).
In our simulated environment, $\Delta a^v = 0.02$ and $\Delta a^{\theta} = 0.2$.
As alluded to in \cref{fg:network-slicc}, it is relatively easy to increase the number of agents in the system---we can simply add new introspective agents in parallel.

Our experiments include two scenarios: $\mathbf{A}^{type}$ + $\mathbf{RP_\alpha}$ and $\mathbf{A}^{type}$ + $\mathbf{RP_\beta}$.
Correspondingly, we use centralized Q-learning with reward $r^g = r_{int} + r^1_{goal} + r^2_{goal} + r^1_{ap} + r^2_{ap}$ as the baseline.
The neural network structure is shown in \cref{fg:network-centralized}.
There is also a single hidden dense layer with $\ell^{central}_{den}$ neurons.
The output is the approximated Q-table which is a matrix with size $\ell^c \times \ell^c$, where $\ell^c = 9$.

\subsection{Training Performance}

\begin{figure}
	\centering
    \includegraphics[width=\linewidth]{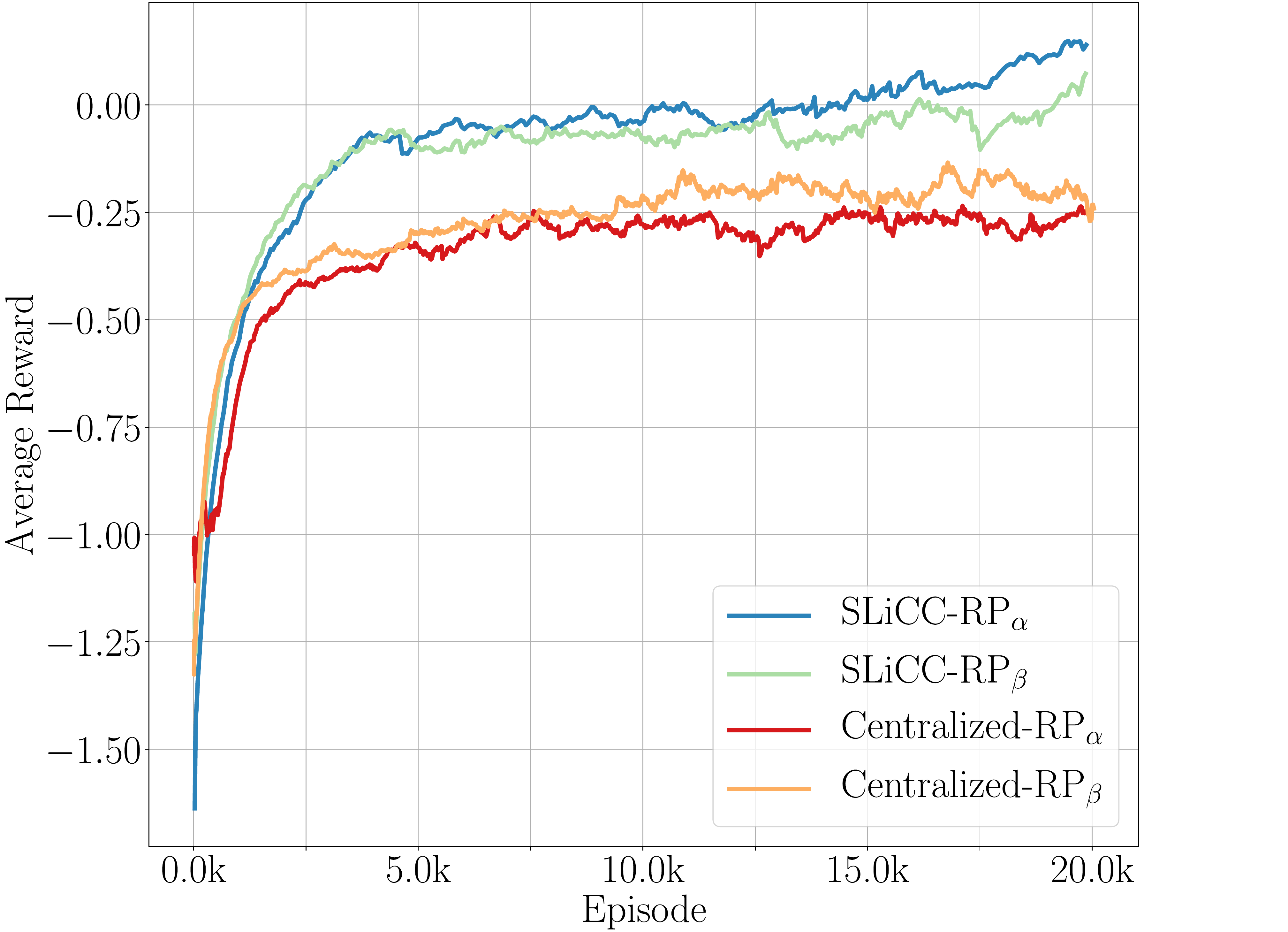} 
    \caption{Average reward per episode for centralized Q-learning and SLiCC with $\mathbf{A}^{type}$. Rewards for SLiCC are expressed as the sum of the rewards of both Agents $P$ and $I$.}\label{fg:learning_curve}
\end{figure}

\begin{figure}
	\centering
    \includegraphics[width=\linewidth]{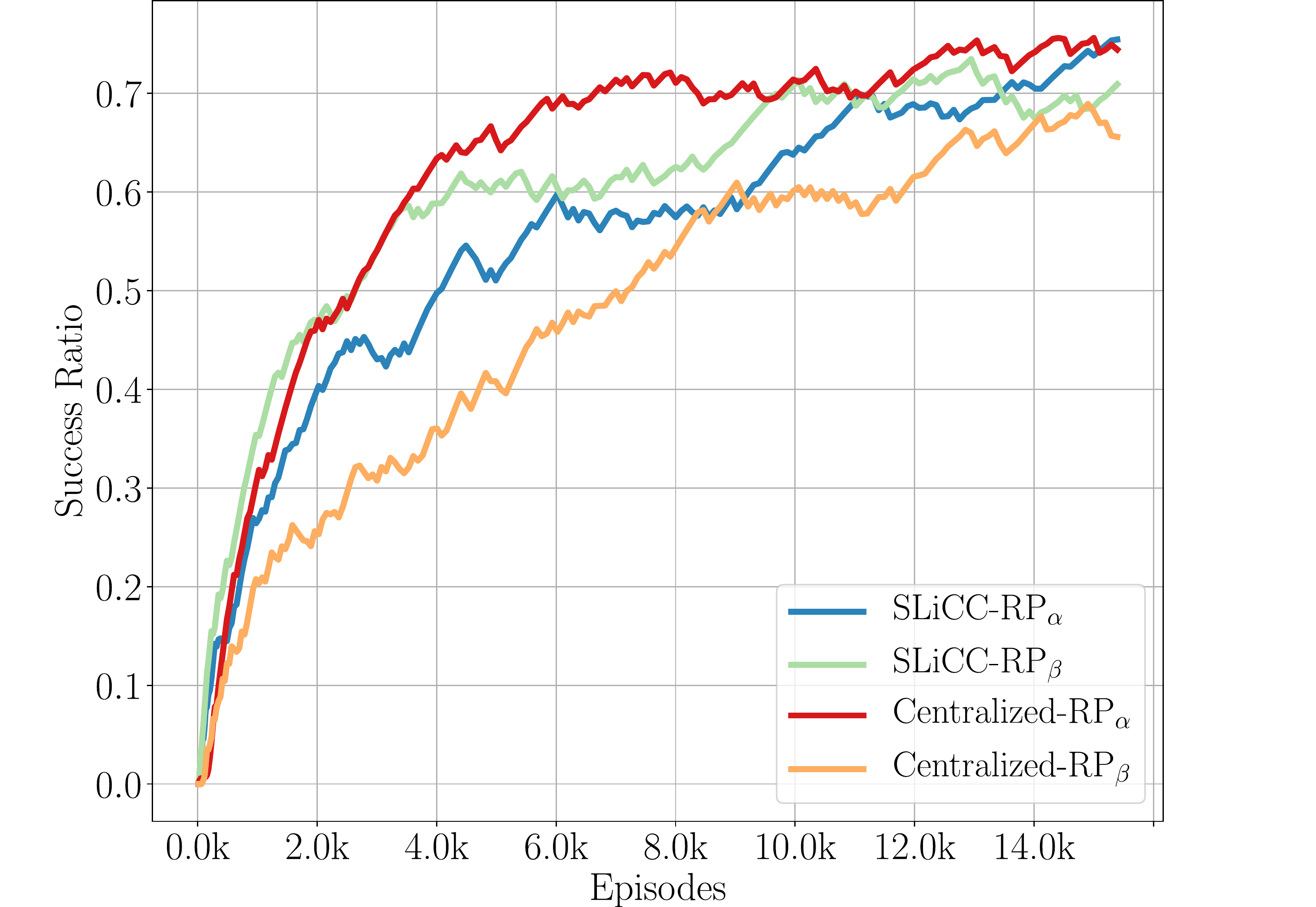} 
    \caption{Success ratio using SLiCC and centralized Q-learning.}\label{fg:success_ratio}
\end{figure}

\cref{fg:learning_curve} shows the average reward per episode attained with SLiCC and centralized Q-learning.
Firstly, SLiCC shows good convergence behavior in our experiments.
This validates our prosocial--introspective cooperation framework, which allows SLiCC to accomplish the common objective with stabilized learned policies.
Despite the system being constrained by partial observations, SLiCC still demonstrates good performance.
Secondly, after around 1,500 episodes, SLiCC has achieved better average reward and maintains this trend until the end of the training process.
We can see that SLiCC takes less time than centralized Q-learning in reaching the same average reward level.
One of the critical reasons that explains why centralized Q-learning has lower average reward is the mutual interference between the three reward constituents: $r_{int}$, $r_{goal}$, and $r_{ap}$.

\cref{fg:success_ratio} shows the success ratio across episodes with SLiCC and centralized Q-learning, for both $\mathbf{RP_\alpha}$ and $\mathbf{RP_\beta}$.
An episode is considered to be successful if three conditions are jointly satisfied:
\begin{enumerate}
    \item the inter-agent distance is close to the desired distance $\sigma$ at all times,
    \item agent velocities are close to the target velocity $(v, \theta)^{target}$ at the end of an episode; and
    \item the episode length exceeds a minimum duration.
\end{enumerate}
As can be seen, SLiCC achieves a good success ratio which is equivalent to that of centralized Q-learning.
While a centralized approach might appear to have an advantage in this experimental scenario because it can command the same action for each robot, this strategy does not work in practice due to noise in transition dynamics.

\subsection{Real Robot Validation}

\begin{figure}
	\centering
    \begin{subfigure}[b]{0.21\textwidth}
    	\centering
        \includegraphics[width=1.0\linewidth]{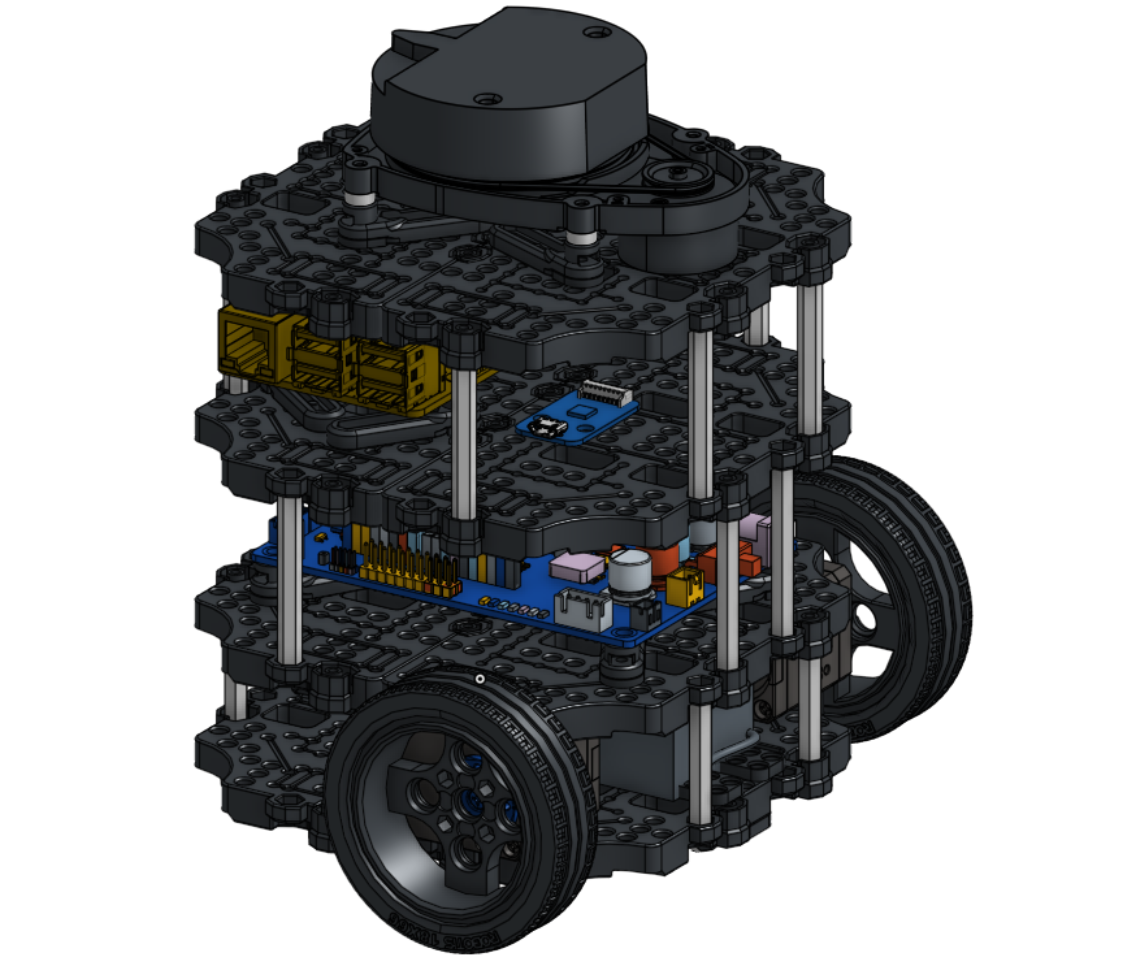}
        \caption{TurtleBot3 Burger mobile robot (CC BY 4.0).}\label{fg:turtlebot3}
    \end{subfigure}
    \hfill
    \begin{subfigure}[b]{0.21\textwidth}
    	\centering
        \includegraphics[width=1.0\linewidth]{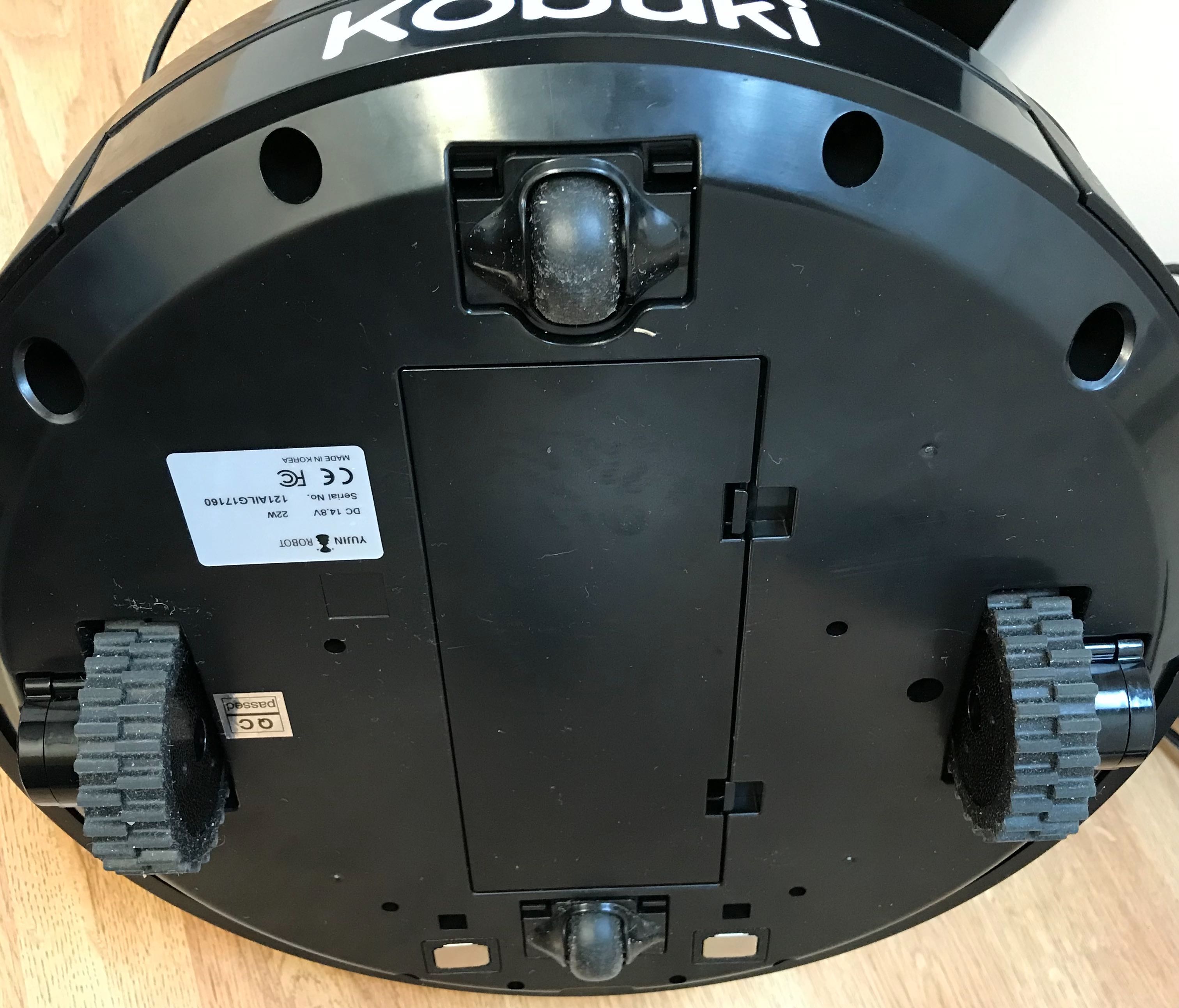}
        \caption{Yujin Robot Kobuki mobile base.}\label{fg:kobuki}
    \end{subfigure}
    \caption{We learn a cooperative control policy in the Gazebo simulation environment with a pair of TurtleBot3 Burger mobile robots (\cref{fg:turtlebot3}). To validate the real-world applicability of the learned policy, we run it on a pair of LoCoBot mobile robots, built on a Yujin Robot Kobuki mobile base (\cref{fg:kobuki}). This introduces different physical dynamics, which confirms the generalizability of learned SLiCC policies. The real robots are able to successfully carry out the object transportation task (\cref{fg:locobot_frames}).}\label{fg:robots}
\end{figure}

\begin{figure*}
	\centering
    \begin{subfigure}[b]{0.19\textwidth}
    	\centering
        \includegraphics[width=1.0\linewidth]{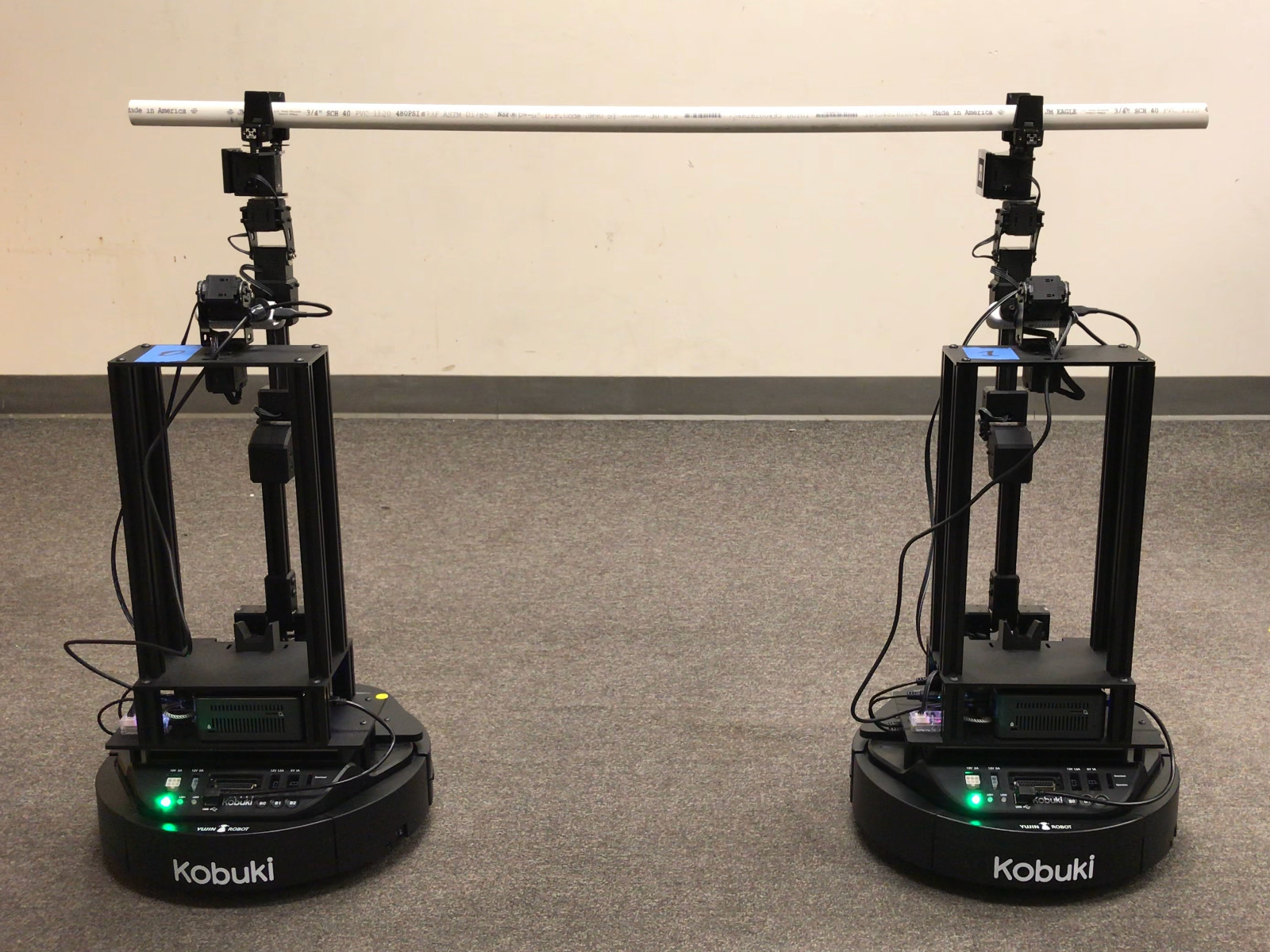}
    \end{subfigure}
    \begin{subfigure}[b]{0.19\textwidth}
    	\centering
        \includegraphics[width=1.0\linewidth]{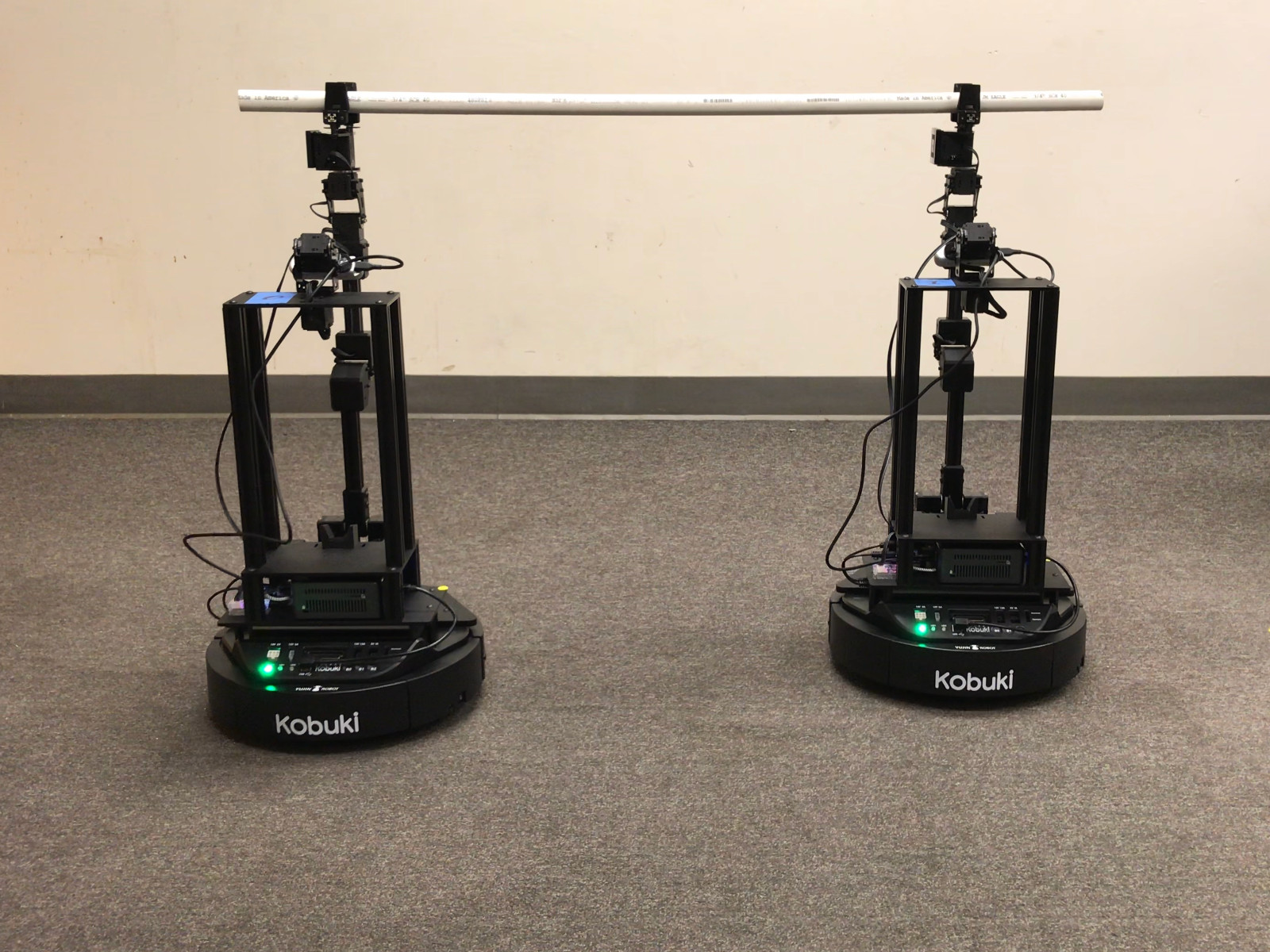}
    \end{subfigure}
    \begin{subfigure}[b]{0.19\textwidth}
    	\centering
        \includegraphics[width=1.0\linewidth]{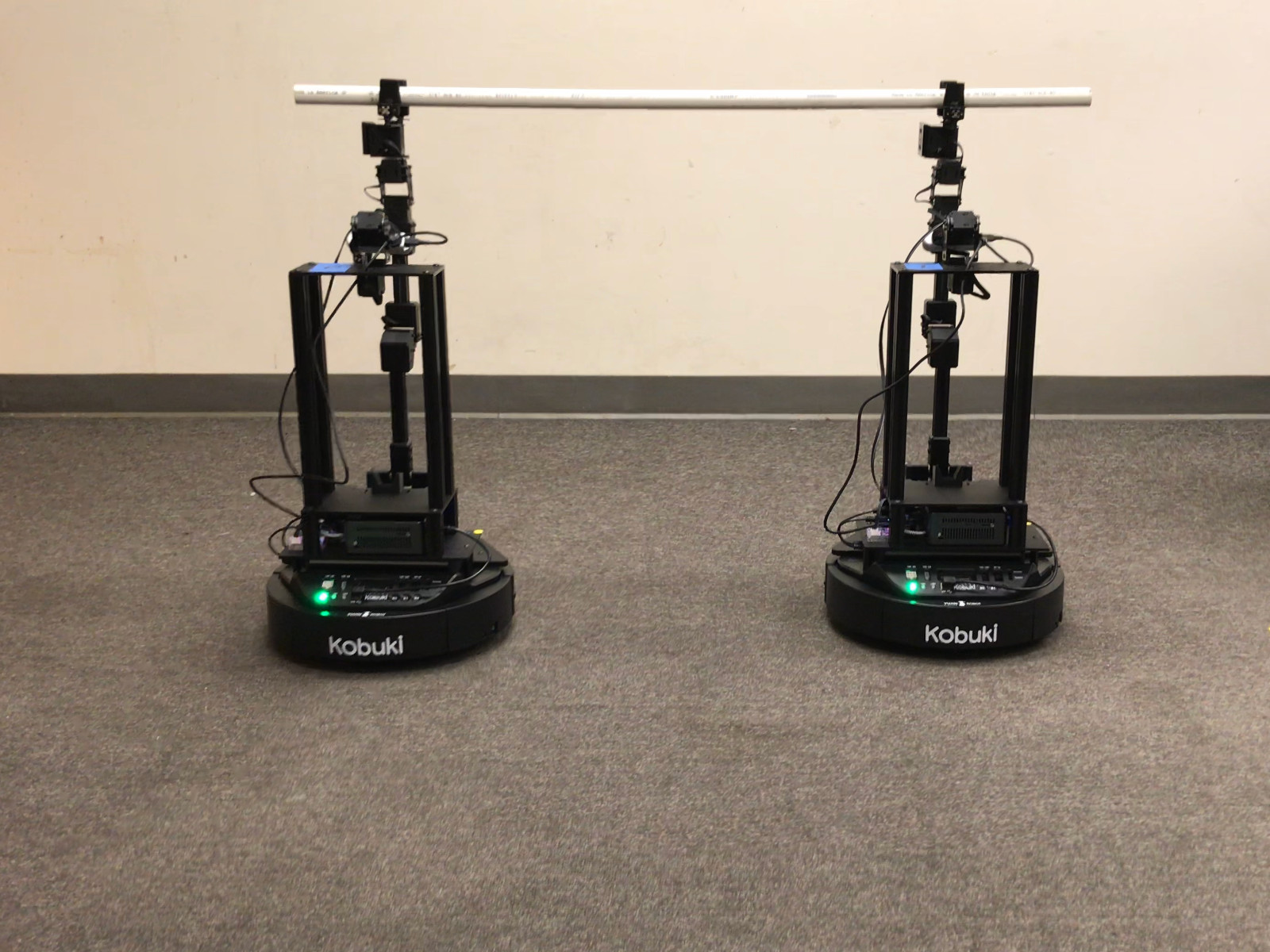}
    \end{subfigure}
    \begin{subfigure}[b]{0.19\textwidth}
    	\centering
        \includegraphics[width=1.0\linewidth]{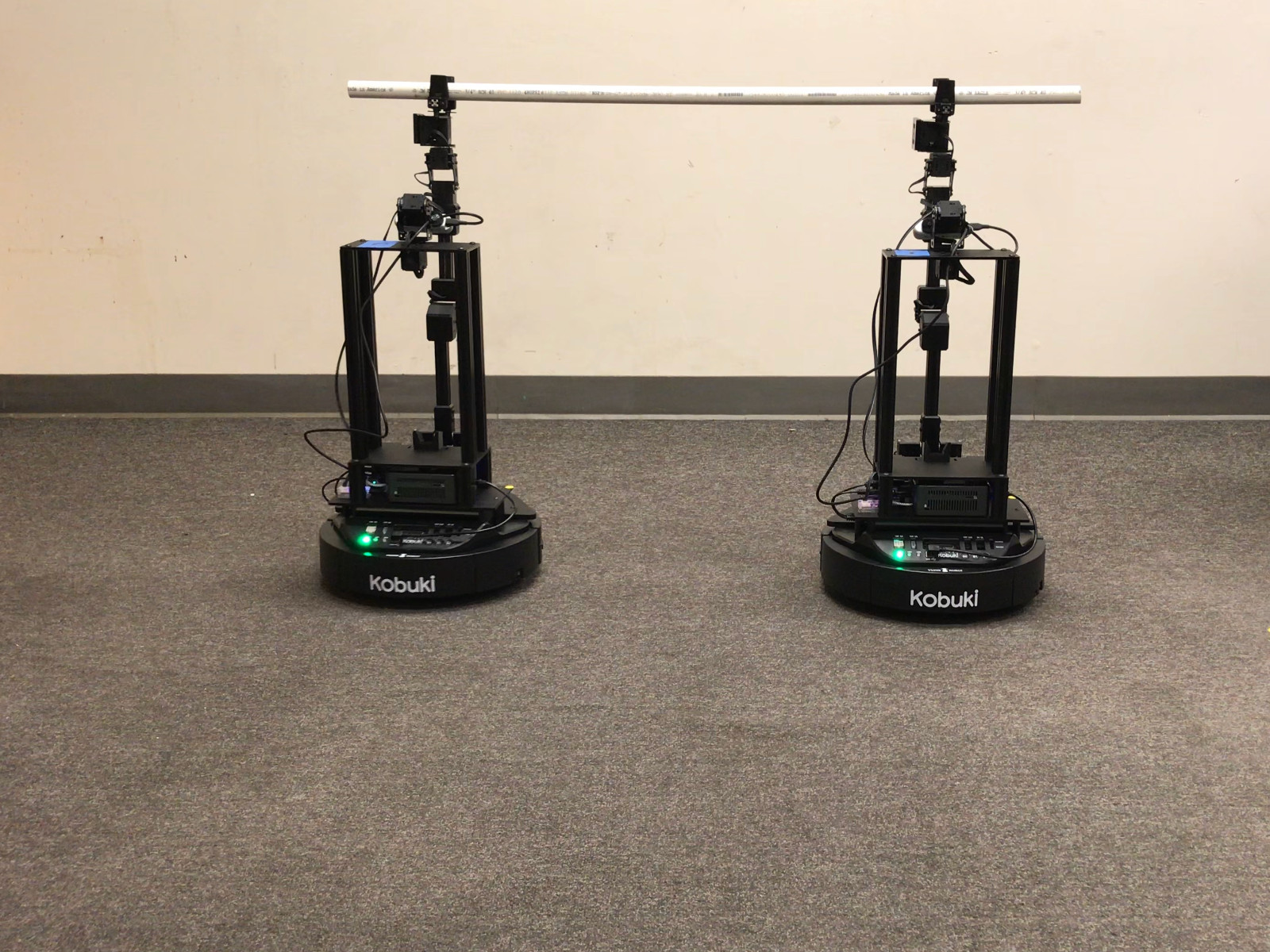}
    \end{subfigure}
    \begin{subfigure}[b]{0.19\textwidth}
    	\centering
        \includegraphics[width=1.0\linewidth]{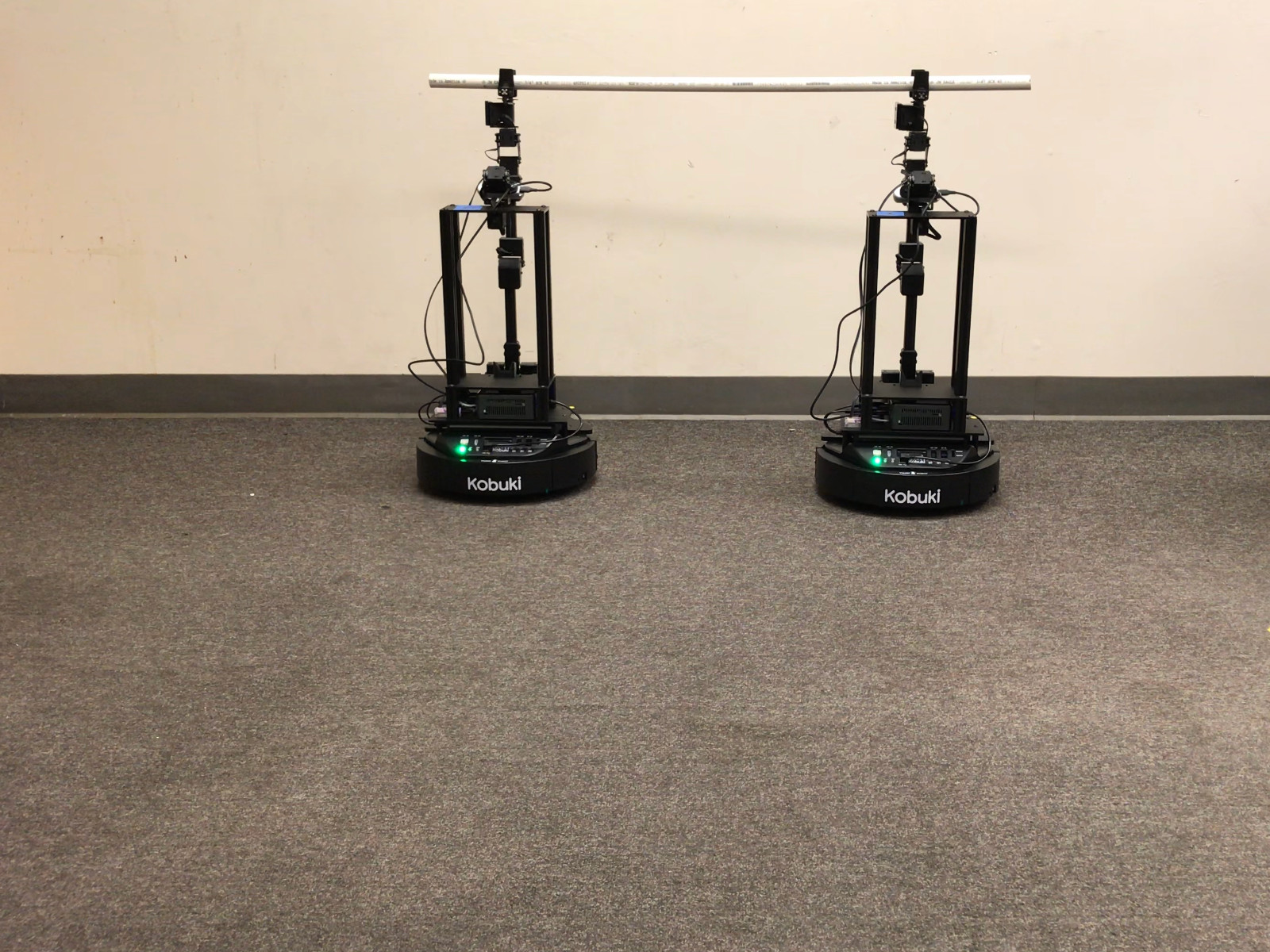}
    \end{subfigure}
    \caption{The pair of real LoCoBot mobile robots carry out the object transportation task using the policies learned on TurtleBot3 Burger mobile robots in simulation. The photo on the far left shows the initial state of the the LoCoBot mobile robots. From left to right, the two robots cooperatively transport the object.}\label{fg:locobot_frames}
\end{figure*}

To validate the real-world applicability of SLiCC, we used the policies learned in simulation to control a pair of real LoCoBot mobile robots.
Notably, the LoCoBot is built on a Yujin Robot Kobuki mobile base (\cref{fg:kobuki}), which has significantly different physical dynamics than the TurtleBot3 (\cref{fg:turtlebot3}).
After scaling the state and action spaces to approximate the spaces used in simulation, the learned policy was able to successfully perform the cooperative transport task without additional training (\cref{fg:locobot_frames}).
This demonstrates the adaptability of SLiCC to different robotic platforms as long as appropriate considerations are made to account for differences in robot dynamics.
These considerations include using generalized dynamics (\cref{eq:dynamics_x,eq:dynamics_y,eq:dynamics_theta,eq:dynamics_v}) and compensating for any magnitude differences in the state and action spaces.

\section{Discussion and Conclusion}\label{sect:conclusion}

In this work, we introduce SLiCC: \underline{S}tackelberg \underline{L}earning \underline{i}n \underline{C}ooperative \underline{C}ontrol, a novel technique to solve the multi-robot cooperation problem in partially observable scenarios.
Our method links state perception with the agents' decision-making strategies through a Stackelberg game--based architecture, which we integrate with partially observable stochastic games (POSG).
Payoff matrices are approximated via deep reinforcement learning, and the proposed framework is evaluated using both simulated and real robots.

In \cref{sect:comparison}, we detailed a number of advantages that SLiCC has over other learning paradigms. In future work, we intend to empirically demonstrate these advantages; some of the potential of the method is still untapped, and we will investigate other scenarios in which we expect SLiCC to perform better (e.g., limited communication bandwidth).
In a separate direction, we also plan to explore alternative training frameworks (e.g., sequential training of the introspective and prosocial agents) and extend SLiCC to continuous action spaces.
Finally, there are several improvements to SLiCC that hold promise for applications in more complex environments (e.g., navigating around obstacles); these will also be pursued in future work.


\section*{Acknowledgment}
We thank the anonymous reviewers and our colleagues for their insightful comments and suggestions which helped improve this manuscript.


\bibliographystyle{IEEEtran}
\bibliography{references}

\end{document}